\documentclass[conference]{IEEEtran}
\IEEEoverridecommandlockouts
\usepackage{cite}
\usepackage{amsmath,amssymb,amsfonts}
\usepackage{algorithmic}
\usepackage{graphicx}
\usepackage{textcomp}
\usepackage{xcolor}
\usepackage[ruled,linesnumbered]{algorithm2e}

\usepackage{subfigure}
\usepackage{multirow}

\makeatletter
\newcommand{\captionof}[1]{\def\@captype{#1}\caption}
\makeatother

\usepackage{tikz}
\usetikzlibrary{arrows.meta,positioning,fit,shapes.geometric}

\def\BibTeX{{\rm B\kern-.05em{\sc i\kern-.025em b}\kern-.08em
    T\kern-.1667em\lower.7ex\hbox{E}\kern-.125emX}}
\begin{document}

\title{An Unbounded Archive-based Transfer Strategy for Dynamic Multi-Objective Optimization with a Changing Number of Objectives\\
}

\author{
Zhiyun Xiao$^{1,2}$, Ke Shang$^{1,2,*}$, Yajun Liu$^{1,2}$, Jianguo Li$^{3}$, Shaojiang Wang$^{3}$, Wei Sun$^{3}$\\[2mm]

$^{1}$School of Artificial Intelligence, Shenzhen University, Shenzhen 518060, China\\
$^{2}$National Engineering Laboratory for Big Data System Computing Technology,\\
Shenzhen University,
Shenzhen 518060, China\\[1mm]
$^{3}$Shenzhen ZTE Software Co., Ltd.
\thanks{* Corresponding author (kshang@foxmail.com)}
}

\maketitle

\begin{abstract}
Dynamic multi-objective optimization with a variable number of objectives is difficult because objective-dimensional variations may significantly change the Pareto front and degrade algorithm adaptability. This paper proposes an unbounded archive-based transfer strategy (UATS), which maintains an unbounded archive of offspring solutions within each environment stage and extracts feasible nondominated solutions as transferable elites when objective changes occur. UATS is embedded into SPEA2SDE to construct UATS-SPEA2SDE, enabling the algorithm to reuse historical evolutionary information while retaining the convergence and diversity advantages of shift-based density estimation. Experiments are conducted on four benchmark problems under three objective-changing settings, where UATS-SPEA2SDE is compared with a restart-based SPEA2SDE baseline and four representative dynamic multi-objective optimization algorithms. The results indicate that the archive-guided transfer improves recovery after environmental changes and enhances adaptability to objective-number variations.
\end{abstract}


\section{Introduction}
Dynamic Multi-Objective Optimization Problems (DMOPs) arise in many time-varying applications, including resource scheduling\cite{Zhou2024CPUGPU}, network optimization\cite{Gao2024IntelligentNetwork}, industrial control, and intelligent transportation systems\cite{Farina2004,Deb2001}. In contrast to static multi-objective problems, a DMOP may involve time-dependent objective functions, constraints, or decision variables. Consequently, the Pareto Front (PF) and Pareto Set (PS) move with the environment, which substantially increases the difficulty of maintaining good solutions\cite{Jiang2017Survey}.

Traditional evolutionary multi-objective optimization algorithms are mainly designed for static environments and usually suffer from performance degradation when the environment changes\cite{Deb2002NSGAII,Zitzler2001SPEA2}. Therefore, many dynamic optimization strategies have been proposed to improve the adaptability and convergence ability of algorithms in changing environments. Existing approaches mainly include diversity enhancement, memory-based strategies, prediction mechanisms, and knowledge transfer methods\cite{Jiang2017Survey,Nguyen2012}.

Among these approaches, archive-based and knowledge transfer mechanisms have shown promising performance. For example, DTAEA adopts a dual-archive mechanism to maintain both convergence and diversity during environmental changes\cite{DTAEA}. KTDMOEA reuses knowledge collected from past environments to guide the search process in the new environment. These studies demonstrate that effectively utilizing historical evolutionary information can significantly improve the tracking ability of algorithms for dynamically changing Pareto Fronts.

SPEA2SDE is a representative multi-objective evolutionary algorithm with good convergence and diversity performance\cite{SPEA2SDE}. However, the original algorithm does not contain a special response component for dynamic cases. When the active objective set changes, SPEA2SDE may adapt slowly because previously generated search information is not explicitly exploited.

Motivated by this limitation, this paper introduces an unbounded archive-based transfer strategy (UATS) for SPEA2SDE. The proposed strategy maintains an external archive without a predefined capacity limit within each environment stage. Offspring solutions generated during the current stage are stored in the archive, and feasible nondominated solutions are extracted as transferable elites when environmental changes occur. Inspired by the idea of knowledge transfer, the archived solutions provide effective guidance for the evolutionary population, enabling the algorithm to converge toward the current Pareto Front more rapidly while maintaining population diversity.

The proposed strategy is evaluated on four DMOP benchmarks in which the number of active objectives increases or decreases over time. It is then incorporated into SPEA2SDE to construct UATS-SPEA2SDE, which is compared with a restart-based SPEA2SDE baseline and four representative dynamic multi-objective optimization algorithms. Experimental results demonstrate that UATS-SPEA2SDE achieves better convergence performance and environmental adaptability under dynamic objective changes.

The main contributions of this paper are summarized as follows:
\begin{itemize}
    \item A UATS mechanism is developed to record offspring during each environmental stage and select nondominated elites for transfer after the active objective set changes.
    \item The proposed UATS is embedded into SPEA2SDE to construct UATS-SPEA2SDE, which improves the adaptability of SPEA2SDE after objective changes.
     \item Four DMOP benchmarks with three dynamic settings are used to compare UATS-SPEA2SDE against a restart baseline and four state-of-the-art algorithms.
\end{itemize}

\section{Related Work}

\subsection{Objective-Number Changes in Dynamic Multi-Objective Optimization}

Multi-objective optimization problems (MOPs) require several conflicting objectives to be optimized at the same time. In practical systems, the search environment is often not fixed and may evolve during the optimization process. DMOPs therefore focus on maintaining solution quality while the PF and PS vary with time.

Among DMOP variants, dynamic multi-objective optimization with a changing number of objectives (DMO-CNO) has become an important topic. Here, the active objective count is allowed to vary, for example when decision needs, user preferences, or evaluation criteria are updated.

Formally, a DMO-CNO problem can be defined as

\begin{equation}
\min_{x \in \Omega} F(x,t)=\left(f_1(x),f_2(x),\ldots,f_{m(t)}(x)\right),
\end{equation}
\noindent
where $x$ is the decision vector, $t$ is the time index, and $m(t)$ gives the objective count at that time. This work considers the dynamic setting in which the active objective number changes, whereas the objective functions are kept fixed over time\cite{Shang2026Benchmark}.

Compared with conventional DMOPs, DMO-CNO problems are more challenging because changes in objective dimensionality may significantly affect the structure of the Pareto Front and population distribution. When objectives are added, the Pareto Front may expand, requiring algorithms to improve diversity and explore new search regions. Conversely, when objectives are removed, the Pareto Front may contract, requiring algorithms to rapidly converge toward a reduced objective space.

To address these challenges, numerous dynamic optimization strategies have been proposed, including diversity enhancement methods, memory-based mechanisms, prediction methods, and transfer learning approaches.

\subsection{Archive and Memory Mechanisms}

Archive and memory mechanisms have been widely adopted in DMOPs to preserve useful historical information during the evolutionary process\cite{Branke1999,Nguyen2012}. By storing high-quality solutions obtained from previous environments, these methods can provide valuable guidance after environmental changes occur and help the population recover convergence more rapidly.

One representative work is DTAEA\cite{DTAEA}, which employs a dual-archive mechanism to simultaneously maintain convergence and diversity. The convergence archive focuses on preserving high-quality solutions close to the Pareto Front, while the diversity archive aims to improve population distribution under changing environments. Such mechanisms have demonstrated strong performance in dynamic many-objective optimization.

Although archive-based approaches can effectively improve adaptability, most existing studies mainly focus on maintaining population diversity or directly reusing historical solutions within a bounded memory structure. When the number of objectives changes, a bounded archive may discard useful historical individuals before they can contribute to later environments. Therefore, how to preserve richer historical search information and transfer it effectively under dynamically changing objective numbers remains an important issue.

\subsection{Representative Algorithms for DMO-CNO}

Recent DMO-CNO studies have produced several representative algorithms. These methods mainly improve environmental adaptability through archive management, knowledge transfer, learning-based estimation, or similarity-based adaptation.

DTAEA (Dynamic Two-Archive Evolutionary Algorithm) is a representative dual-archive optimization method. By managing convergence-oriented and diversity-oriented archives in parallel, it can keep useful search directions and population spread after environmental changes, making it a widely studied dynamic multi-objective optimizer.

KTDMOEA (Knowledge Transfer Dynamic Multi-objective Evolutionary Algorithm) is a knowledge transfer-based dynamic multi-objective optimization algorithm\cite{KTDMOEA,TransferSurvey}. The algorithm utilizes valuable population information obtained from historical environments to guide the search process in newly changed environments, thereby improving adaptability and Pareto Front tracking performance.

LEC (Learning to Expand and Contract PSs) represents a class of learning-based dynamic optimization methods\cite{LEC}. By learning evolutionary patterns from historical environments, LEC estimates the potential distribution of Pareto-optimal solutions in new environments, thereby improving search efficiency and environmental adaptability.

STA (Similarity Transfer Approach) is a dynamic optimization approach that combines spatial and temporal information\cite{STA}. The method utilizes temporal relationships between environments and spatial distribution characteristics of populations to dynamically adjust the population after environmental changes, improving convergence performance and stability in complex dynamic environments.

SPEA2SDE is a representative many-objective evolutionary algorithm based on Shift-based Density Estimation (SDE). By improving density estimation in high-dimensional objective spaces, SPEA2SDE effectively balances convergence and diversity. However, SPEA2SDE is mainly designed for static optimization problems, and its capability of utilizing historical information in dynamic environments with changing objective numbers remains limited.

Although these algorithms have shown promising performance, existing archive-based and transfer-based methods may still rely on bounded memory structures, relatively complex learning models, or assumptions about the similarity between consecutive environments. Therefore, this paper proposes an unbounded archive-based transfer strategy (UATS) to preserve richer historical elite solutions and reuse them after objective changes. In the experimental study, UATS is instantiated with SPEA2SDE to construct UATS-SPEA2SDE.

\subsection{Challenges of DMO-CNO}

Compared with conventional dynamic multi-objective optimization problems, DMO-CNO introduces additional challenges because objective changes directly modify the dimensionality and structure of the objective space\cite{DTAEA,KTDMOEA}. When new objectives are introduced, the Pareto Front (PF) may expand and previously well-converged solutions can become suboptimal. Conversely, when objectives are removed, some previously dominated solutions may become non-dominated, leading to significant changes in population distribution\cite{DTAEA}.

Another challenge is that the usefulness of historical knowledge becomes uncertain after objective changes. Many memory-based and transfer-learning approaches assume that consecutive environments share similar PF characteristics. However, changes in objective dimensionality may substantially alter the optimization landscape, reducing the reliability of previously acquired knowledge\cite{KTDMOEA,TransferSurvey}. Therefore, transfer strategies for DMO-CNO should preserve useful historical information while avoiding excessive dependence on environment similarity.

In addition, maintaining a balance between convergence and diversity becomes more difficult when the number of objectives changes. Objective increases generally require stronger exploration to cover expanded search regions, whereas objective decreases demand rapid convergence toward the reduced objective space\cite{DTAEA,KTDMOEA}. When the number of objectives becomes large, many-objective optimization difficulties further arise, including weaker dominance relationships and reduced selection pressure\cite{SPEA2SDE,Xue2022E3A}.

\section{Proposed Method}

\subsection{Overview of the Proposed Transfer Strategy}

This paper proposes UATS to improve the response of evolutionary algorithms on DMO-CNO problems. UATS is designed as a general framework that can be incorporated into different static evolutionary algorithms. The key idea is to maintain an unbounded external archive within each environment stage, store offspring solutions generated by the base algorithm, and extract feasible nondominated solutions as transferable elites when environmental changes occur.

The core idea of UATS is inspired by knowledge transfer. Instead of discarding all historical information after environmental changes, the base algorithm uses transferable elite solutions extracted from the unbounded archive to guide the search process in the new environment. In this way, the population can rapidly adapt to the changed objective space and converge toward the current Pareto Front (PF).

The framework contains three main components:

\begin{itemize}
    \item Environmental change detection;
    \item Unbounded archive-based solution transfer;
    \item Population reconstruction after environmental changes.
\end{itemize}

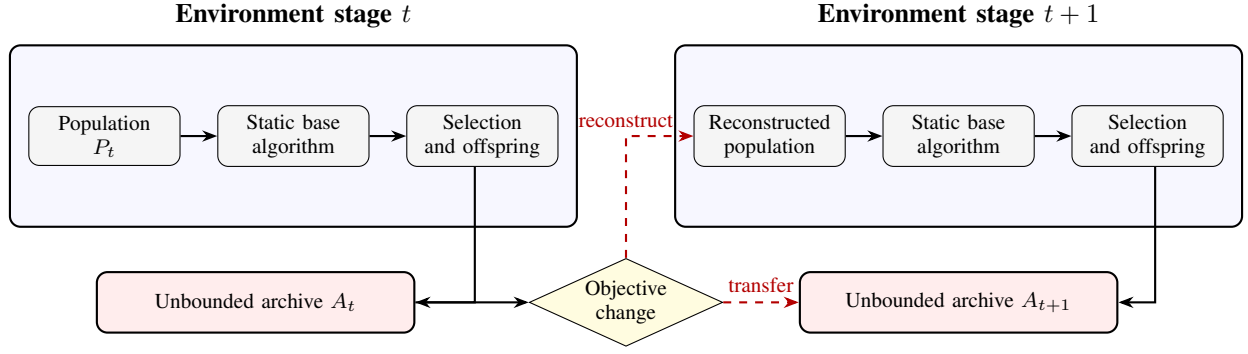
\begin{figure*}[!htb]
\centering
\begin{tikzpicture}[
    font=\footnotesize,
    node distance=7mm,
    stagebox/.style={draw, rounded corners, thick, align=center, minimum width=75mm, minimum height=24mm, fill=blue!3},
    block/.style={draw, rounded corners, align=center, minimum width=20mm, minimum height=7mm, fill=gray!8},
    archive/.style={draw, rounded corners, thick, align=center, minimum width=42mm, minimum height=8mm, fill=red!7},
    detect/.style={draw, diamond, aspect=2.2, align=center, inner sep=1pt, fill=yellow!15},
    arrow/.style={-{Stealth[length=2mm]}, thick},
    transfer/.style={-{Stealth[length=2mm]}, thick, dashed, red!70!black}
]

\node[stagebox] (stage1) at (0,0) {};
\node[stagebox] (stage2) at (88mm,0) {};
\node[font=\bfseries] at (stage1.north) [above=1mm] {Environment stage $t$};
\node[font=\bfseries] at (stage2.north) [above=1mm] {Environment stage $t+1$};

\node[block] (p1) at ([xshift=-25mm]stage1.center) {Population\\$P_t$};
\node[block] (b1) at (stage1.center) {Static base\\algorithm};
\node[block] (s1) at ([xshift=25mm]stage1.center) {Selection\\and offspring};

\node[block] (p2) at ([xshift=-25mm]stage2.center) {Reconstructed\\population};
\node[block] (b2) at (stage2.center) {Static base\\algorithm};
\node[block] (s2) at ([xshift=25mm]stage2.center) {Selection\\and offspring};

\node[archive] (a1) at (-5mm,-22mm) {Unbounded archive $A_t$};
\node[archive] (a2) at (88mm,-22mm) {Unbounded archive $A_{t+1}$};
\node[detect] (d) at (44mm,-22mm) {Objective\\change};

\draw[arrow] (p1) -- (b1);
\draw[arrow] (b1) -- (s1);
\draw[arrow] (p2) -- (b2);
\draw[arrow] (b2) -- (s2);

\draw[arrow] ([xshift=-1mm]s1.south) |- (a1.east);
\draw[arrow] ([xshift=1mm]s2.south) |- (a2.east);

\draw[arrow] (a1) -- (d);
\draw[transfer] (d) -- node[above]{transfer} (a2);
\draw[transfer] (d.north) |- (p2.west);
\node[red!70!black, anchor=west] at ([xshift=-7.5mm, yshift=2mm]d.north |- p2.west) {reconstruct};

\end{tikzpicture}
\caption{Overall framework of UATS across two consecutive environment stages.}
\label{fig:uats_framework}
\end{figure*}

During each static evolutionary stage, offspring solutions generated by the base algorithm are continuously stored in the unbounded archive. Unlike bounded archive mechanisms that restrict the archive size in advance, UATS does not impose a fixed archive boundary, so richer evolutionary information can be retained within each environment stage before the next environmental change occurs. Once an environmental change is detected, transferable nondominated elite solutions are extracted and transferred into the next environment to reconstruct the population. It should be noted that UATS does not always use direct projection after objective changes. Direct projection is adopted only when the new active objective subset is contained in the previous one. Otherwise, the transferred decision variables are re-evaluated under the new active objective subset. The overall framework of UATS is illustrated in Fig.~\ref{fig:uats_framework}.

\begin{algorithm}[t]
\caption{General UATS for DMO-CNO}
\label{alg:framework}

Initialize unbounded archive $A$\;

\While{termination criterion is not satisfied}{

    Detect environmental changes\;

    \If{environment changes}{

        Extract feasible nondominated solutions from archive $A$ as transferable elites\;

        \If{$I_{\mathrm{new}}\subseteq I_{\mathrm{old}}$}{
            Project transferred solutions onto the new objective subset\;
        }
        \Else{
            Re-evaluate transferred decision variables under $I_{\mathrm{new}}$\;
        }
        Reconstruct population using transferred solutions\;
    }

    Execute evolutionary search using the base algorithm\;

    Update unbounded archive $A$\;

}

\end{algorithm}

\subsection{Environmental Change Detection}

In UATS, environmental changes are detected at the beginning of each evolutionary generation. The current environment stage is represented by a stage signature containing the current objective configuration.

Let

\begin{equation}
S(t)=\{m(t),I(t)\},
\end{equation}
\noindent
where $m(t)$ denotes the number of objectives at time step $t$, and $I(t)$ represents the active objective subset.

At each generation, the algorithm compares the current stage signature with that of the previous generation. If the two signatures are different, an environmental change is considered to have occurred.

Compared with traditional passive adaptation strategies, the proposed method performs environment detection before offspring generation, ensuring that the population structure is always synchronized with the current environment.

\subsection{Unbounded Archive-Based Solution Transfer}

To improve adaptation ability after environmental changes, UATS maintains an unbounded external archive during each environment stage. In the proposed implementation, the archive stores offspring solutions generated by the base algorithm. The term ``unbounded'' means that no predefined archive capacity or truncation operation is imposed within one environment stage.

When an environmental change occurs, feasible nondominated solutions in the archive are extracted and regarded as transferable elite solutions. If the archive is empty, feasible nondominated solutions from the current population are used instead. These selected solutions are then transferred into the new environment for population reconstruction.

Although the archive is unbounded within each environment stage, it is reset after an environmental change and population reconstruction. Therefore, the archive does not indefinitely accumulate solutions over all environmental stages. Assuming that $N$ offspring solutions are generated at each generation and one environment lasts for $\tau_t$ generations, the archive size within one stage is at most $O(\tau_t N)$. The corresponding storage cost is $O(\tau_t N D)$, where $D$ is the number of decision variables. The archive update only requires appending offspring solutions, resulting in an additional maintenance cost of $O(N)$ per generation.

Two different transfer strategies are designed according to the type of objective change.

\subsubsection{Objective Increase}

When the number of objectives increases, the dimensionality of the objective space expands, causing previously optimal solutions to become partially insufficient for the new environment. In this case, transferred solutions are re-evaluated under the new objective configuration:

\begin{equation}
\mathbf{f}'(x)=\left(f_1(x),f_2(x),\ldots,f_{m(t+1)}(x)\right).
\end{equation}

The re-evaluated solutions provide useful search guidance while allowing the population to gradually adapt to newly introduced objectives.

\subsubsection{Objective Decrease}

When the number of objectives decreases, part of the objective dimensions may be removed. However, objective reduction does not necessarily mean that the new active objective subset is fully contained in the previous one. In severe dynamic settings, the number of objectives may decrease while some newly activated objectives appear in the new subset.

Let $I_{\mathrm{old}}$ and $I_{\mathrm{new}}$ denote the active objective subsets before and after an environmental change, respectively. If $I_{\mathrm{new}}\subseteq I_{\mathrm{old}}$, the objective vectors of transferred solutions are directly projected onto the remaining objective dimensions:
\begin{equation}
F_{new}(x)=\Pi_{I_{\mathrm{new}}}(F_{old}(x)).
\end{equation}

Otherwise, direct projection is invalid because some objective values required in the new environment are unavailable. In this case, UATS re-evaluates the transferred decision variables under the new active objective subset:
\begin{equation}
F_{new}(x)=\left(f_i(x)\right)_{i\in I_{\mathrm{new}}}.
\end{equation}
Thus, direct projection is used only for nested objective-reduction cases, while re-evaluation is adopted for non-nested objective changes.

\subsection{Population Reconstruction Mechanism}

After the transfer process, the population is reconstructed to match the required population size.

If the number of transferred elite solutions exceeds the predefined population size, a subset of transferred solutions is randomly selected. Otherwise, randomly initialized individuals are introduced to supplement the population:

\begin{equation}
P_{new}=P_{transfer}\cup P_{random}.
\end{equation}

This reconstruction strategy enables UATS to preserve historical convergence information while maintaining sufficient population diversity in the new environment.

\subsection{Integration with SPEA2SDE}

As UATS is a general transfer framework, it can be combined with different evolutionary multi-objective optimization algorithms.SPEA2SDE is adopted as the base algorithm to instantiate the proposed framework, resulting in UATS-SPEA2SDE.

After population reconstruction, UATS-SPEA2SDE continues the evolutionary process using the standard SPEA2SDE environmental selection and shift-based density estimation mechanisms. Specifically, tournament selection is first applied to generate the mating pool. Offspring solutions are then generated through genetic operators. Finally, environmental selection based on SPEA2SDE fitness evaluation is performed to maintain convergence and diversity simultaneously.

By integrating UATS with SPEA2SDE, the instantiated algorithm effectively utilizes historical evolutionary information while preserving the convergence and diversity advantages of SPEA2SDE in dynamic environments with changing objective numbers.

% Queue the three result groups early enough for pages 6 and 7. Do not force a
% page break here: the preceding text should continue into the right column.
\setlength{\dblfloatsep}{4pt}
\setlength{\dbltextfloatsep}{4pt}
\begin{figure*}[!t]
\centering
\captionof{table}{MHV results (mean $\pm$ standard deviation) under \textbf{Setting I}.Bold values indicate the best result in each row.}
\label{tab:mhv_results_set1}

\fontsize{7.5pt}{9pt}\selectfont
\setlength{\tabcolsep}{5.75pt}

\begin{tabular}{c c c c c c c c}
\hline
Problem & $\tau_t$
& DTAEA & KTDMOEA & LEC & STA & R-SPEA2SDE & UATS-SPEA2SDE \\
\hline

\multirow{3}{*}{Minus-DTLZ1}
& 25
& 0.5310 (3.15e-03)
& 0.5299 (5.24e-03)
& 0.5010 (1.85e-02)
& 0.4216 (9.46e-03)
& 0.5412 (3.24e-03)
& \textbf{0.6544 (1.53e-02)}\\
& 50
& 0.5300 (1.61e-03)
& 0.5312 (2.52e-03)
& 0.5411 (5.69e-03)
& 0.4809 (6.75e-03)
& 0.6173 (3.55e-03)
& \textbf{0.6988 (4.62e-03)}\\
& 100
& 0.5249 (6.68e-04)
& 0.5256 (7.91e-04)
& 0.5297 (1.77e-03)
& 0.5118 (3.26e-03)
& 0.6795 (3.28e-03)
& \textbf{0.7066 (1.11e-03)}\\
\hline

\multirow{3}{*}{Minus-DTLZ2}
& 25
& 0.6286 (4.26e-03)
& 0.6238 (7.27e-03)
& 0.6007 (2.37e-02)
& 0.5908 (6.80e-03)
& 0.7417 (2.56e-03)
& \textbf{0.7738 (5.52e-03)}\\
& 50
& 0.6169 (2.19e-03)
& 0.6167 (3.06e-03)
& 0.6407 (7.05e-03)
& 0.6215 (2.44e-03)
& 0.7797 (8.17e-04)
& \textbf{0.7891 (8.60e-04)}\\
& 100
& 0.6082 (1.37e-03)
& 0.6077 (1.44e-03)
& 0.6160 (2.83e-03)
& 0.6104 (1.81e-03)
& 0.7855 (8.83e-04)
& \textbf{0.7896 (7.57e-04)}\\
\hline

\multirow{3}{*}{Minus-DTLZ3}
& 25
& 0.6290 (3.67e-03)
& 0.6281 (7.59e-03)
& 0.5962 (2.01e-02)
& 0.5319 (1.12e-02)
& 0.6121 (3.77e-03)
& \textbf{0.7645 (6.83e-03)}\\
& 50
& 0.6200 (1.97e-03)
& 0.6221 (2.65e-03)
& 0.6403 (1.05e-02)
& 0.5886 (5.48e-03)
& 0.6971 (4.69e-03)
& \textbf{0.7879 (1.07e-03)}\\
& 100
& 0.6105 (1.31e-03)
& 0.6106 (1.85e-03)
& 0.6195 (3.54e-03)
& 0.6049 (2.80e-03)
& 0.7667 (3.63e-03)
& \textbf{0.7894 (7.79e-04)}\\
\hline

\multirow{3}{*}{Minus-DTLZ4}
& 25
& 0.6223 (5.84e-03)
& 0.6029 (2.73e-02)
& 0.3310 (1.23e-01)
& 0.5678 (2.24e-02)
& 0.7098 (3.06e-03)
& \textbf{0.7273 (1.55e-01)}\\
& 50
& 0.6201 (3.15e-03)
& 0.6153 (1.79e-02)
& 0.4310 (1.13e-01)
& 0.6193 (8.32e-03)
& 0.7783 (8.63e-04)
& \textbf{0.7850 (2.50e-02)}\\
& 100
& 0.6083 (1.96e-03)
& 0.6076 (4.28e-03)
& 0.5429 (6.26e-02)
& 0.6080 (1.11e-02)
& 0.7851 (8.92e-04)
& \textbf{0.7887 (6.50e-03)}\\
\hline
% Continue remaining benchmark problems...

\end{tabular}

\makeatletter
\def\@captype{figure}
\makeatother

\subfigure[$\tau_t=25$]{
    \includegraphics[width=0.31\textwidth]{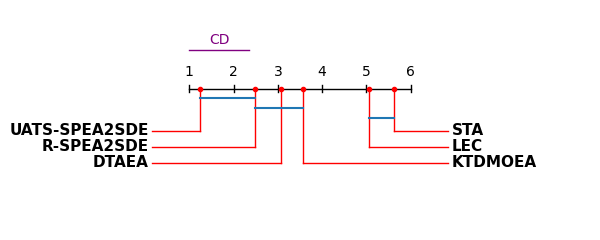}
    \label{fig:setting1_tau25}
}
\hfill
\subfigure[$\tau_t=50$]{
    \includegraphics[width=0.31\textwidth]{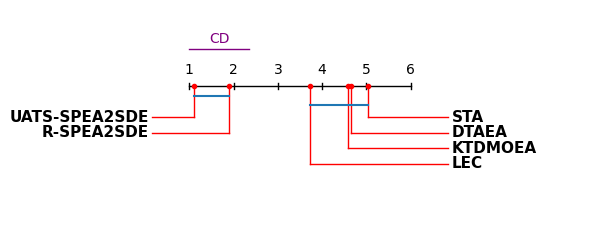}
    \label{fig:setting1_tau50}
}
\hfill
\subfigure[$\tau_t=100$]{
    \includegraphics[width=0.31\textwidth]{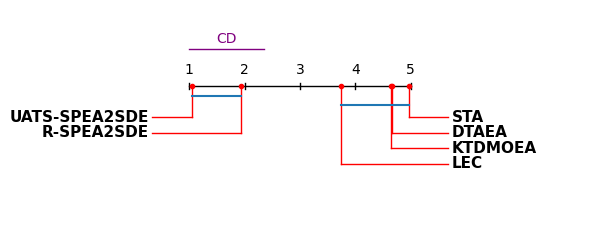}
    \label{fig:setting1_tau100}
}

\caption{Friedman ranking under \textbf{Setting I}. A smaller rank means better performance.}

\label{fig:mhv_setting1}

\end{figure*}
\begin{figure*}[!t]
\centering
\captionof{table}{MHV results (mean $\pm$ standard deviation) under \textbf{Setting II}.Bold values indicate the best result in each row.}
\label{tab:mhv_results_set2}

\fontsize{7.5pt}{9pt}\selectfont
\setlength{\tabcolsep}{5.75pt}

\begin{tabular}{c c c c c c c c}
\hline
Problem & $\tau_t$
& DTAEA & KTDMOEA & LEC & STA & R-SPEA2SDE & UATS-SPEA2SDE \\
\hline

\multirow{3}{*}{Minus-DTLZ1}
& 25
& 0.3005 (3.47e-03)
& 0.3166 (5.76e-03)
& 0.3110 (1.48e-02)
& 0.2527 (5.40e-03)
& 0.3128 (2.88e-03)
& \textbf{0.4088 (7.49e-03)}\\
& 50
& 0.3177 (1.67e-03)
& 0.3171 (5.06e-03)
& 0.3258 (5.83e-03)
& 0.2903 (4.30e-03)
& 0.3794 (2.87e-03)
& \textbf{0.4373 (4.37e-03)}\\
& 100
& 0.3200 (1.50e-03)
& 0.3180 (4.50e-03)
& 0.3300 (5.00e-03)
& 0.3100 (3.50e-03)
& 0.4244 (2.83e-03)
& \textbf{0.4476 (2.47e-03)}\\
\hline

\multirow{3}{*}{Minus-DTLZ2}
& 25
& 0.4062 (2.75e-03)
& 0.4068 (1.02e-02)
& 0.3845 (1.72e-02)
& 0.3694 (8.03e-03)
& 0.4744 (3.45e-03)
& \textbf{0.5155 (6.53e-03)} \\
& 50
& 0.4047 (2.19e-03)
& 0.4063 (3.27e-03)
& 0.4223 (4.80e-03)
& 0.4064 (2.56e-03)
& 0.5242 (1.27e-03)
& \textbf{0.5371 (9.57e-04)}\\
& 100
& 0.4040 (2.00e-03)
& 0.4060 (3.00e-03)
& 0.4300 (4.00e-03)
& 0.4200 (2.50e-03)
& 0.5335 (1.21e-03)
& \textbf{0.5383 (6.14e-04)}\\
\hline

\multirow{3}{*}{Minus-DTLZ3}
& 25
& 0.3947 (4.87e-03)
& 0.4086 (7.76e-03)
& 0.3697 (1.79e-02)
& 0.3227 (8.60e-03)
& 0.3867 (3.19e-03)
& \textbf{0.5064 (6.62e-03)}\\
& 50
& 0.4051 (2.58e-03)
& 0.4110 (2.63e-03)
& 0.4197 (5.71e-03)
& 0.3708 (8.18e-03)
& 0.4582 (4.18e-03)
& \textbf{0.5357 (1.38e-03)}\\
& 100
& 0.4100 (2.50e-03)
& 0.4120 (2.50e-03)
& 0.4300 (4.50e-03)
& 0.3900 (4.00e-03)
& 0.5135 (3.93e-03)
& \textbf{0.5383 (8.25e-04)}\\
\hline

\multirow{3}{*}{Minus-DTLZ4}
& 25
& 0.3882 (3.44e-03)
& 0.3619 (2.98e-02)
& 0.2235 (6.86e-02)
& 0.3354 (1.63e-02)
& 0.4515 (3.72e-03)
& \textbf{0.4850 (8.50e-03)}\\
& 50
& 0.4033 (1.79e-03)
& 0.3915 (8.72e-03)
& 0.2873 (4.54e-02)
& 0.3947 (4.27e-03)
& 0.5186 (1.02e-03)
& \textbf{0.5350 (3.50e-03)}\\
& 100
& 0.4100 (1.50e-03)
& 0.4050 (6.50e-03)
& 0.3200 (2.50e-02)
& 0.4200 (3.50e-03)
& 0.5277 (8.35e-04)
& \textbf{0.5420 (1.20e-03)}\\
\hline
% Continue remaining benchmark problems...

\end{tabular}

\makeatletter
\def\@captype{figure}
\makeatother

\subfigure[$\tau_t=25$]{
    \includegraphics[width=0.31\textwidth]{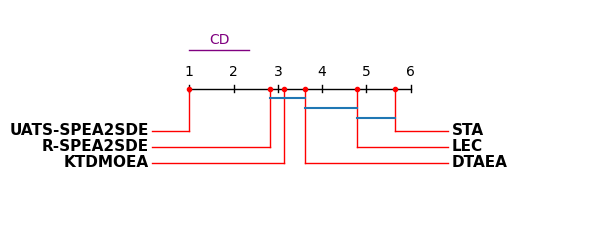}
    \label{fig:setting2_tau25}
}
\hfill
\subfigure[$\tau_t=50$]{
    \includegraphics[width=0.31\textwidth]{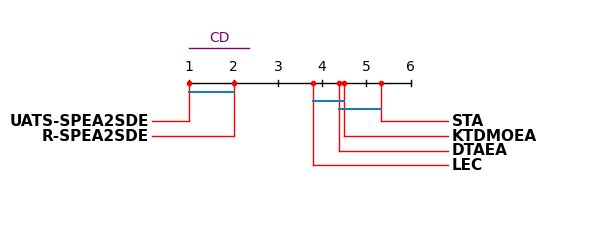}
    \label{fig:setting2_tau50}
}
\hfill
\subfigure[$\tau_t=100$]{
    \includegraphics[width=0.31\textwidth]{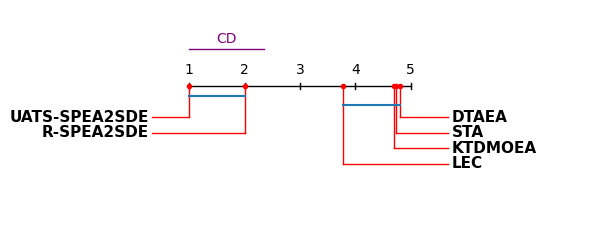}
    \label{fig:setting2_tau100}
}

\caption{Friedman ranking under \textbf{Setting II}. A smaller rank means better performance.}

\label{fig:mhv_setting2}

\end{figure*}
\begin{figure*}[!t]
\centering
\captionof{table}{MHV results (mean $\pm$ standard deviation) under \textbf{Setting III}.Bold values indicate the best result in each row.}
\label{tab:mhv_results_set3}

\fontsize{7.5pt}{9pt}\selectfont
\setlength{\tabcolsep}{5.75pt}

\begin{tabular}{c c c c c c c c}
\hline
Problem & $\tau_t$ 
& DTAEA & KTDMOEA & LEC & STA & R-SPEA2SDE & UATS-SPEA2SDE \\
\hline

\multirow{3}{*}{Minus-DTLZ1} 
& 25  
& 0.2512 (3.11e-03) 
& 0.2590 (4.20e-03) 
& 0.2455 (8.50e-03) 
& 0.2105 (6.40e-03) 
& 0.2650 (3.83e-03) 
& \textbf{0.3066 (1.37e-02)}\\
& 50  
& 0.2688 (2.45e-03) 
& 0.2705 (3.80e-03) 
& 0.2580 (6.10e-03) 
& 0.2310 (4.20e-03) 
& 0.3246 (3.53e-03) 
& \textbf{0.3550 (5.20e-03)}\\
& 100 
& 0.2745 (1.89e-03) 
& 0.2750 (2.50e-03) 
& 0.2630 (5.00e-03) 
& 0.2450 (3.10e-03) 
& 0.3682 (3.44e-03) 
& \textbf{0.3850 (3.10e-03)}\\
\hline

\multirow{3}{*}{Minus-DTLZ2} 
& 25  
& 0.3580 (3.20e-03) 
& 0.3610 (5.50e-03) 
& 0.3340 (1.20e-02) 
& 0.3150 (6.80e-03) 
& 0.4213 (2.30e-03) 
& \textbf{0.4455 (7.89e-03)} \\
& 50  
& 0.3625 (2.10e-03) 
& 0.3680 (3.40e-03) 
& 0.3510 (8.50e-03) 
& 0.3280 (4.50e-03) 
& 0.4726 (9.18e-04) 
& \textbf{0.4835 (1.18e-03)}\\
& 100 
& 0.3650 (1.80e-03) 
& 0.3705 (2.20e-03) 
& 0.3610 (5.40e-03) 
& 0.3350 (3.20e-03) 
& 0.4815 (6.25e-04) 
& \textbf{0.4869 (8.57e-04)}\\
\hline

\multirow{3}{*}{Minus-DTLZ3} 
& 25  
& 0.3450 (4.50e-03)
& 0.3480 (6.20e-03) 
& 0.3050 (1.50e-02) 
& 0.2850 (7.50e-03) 
& 0.3367 (2.85e-03) 
& \textbf{0.4250 (5.80e-03)}\\
& 50  
& 0.3550 (3.10e-03) 
& 0.3610 (4.80e-03) 
& 0.3250 (1.10e-02) 
& 0.2980 (5.20e-03)  
& 0.4047 (4.08e-03) 
& \textbf{0.4650 (3.50e-03)}\\
& 100 
& 0.3620 (2.50e-03) 
& 0.3680 (3.50e-03) 
& 0.3380 (8.50e-03) 
& 0.3120 (4.10e-03) 
& 0.4608 (3.16e-03) 
& \textbf{0.4850 (2.10e-03)}\\
\hline

\multirow{3}{*}{Minus-DTLZ4} 
& 25  
& 0.3380 (3.80e-03) 
& 0.3150 (8.50e-03) 
& 0.1950 (4.50e-02) 
& 0.2950 (1.10e-02) 
& 0.3991 (4.23e-03) 
& \textbf{0.4150 (1.80e-02)}\\
& 50  
& 0.3450 (2.50e-03) 
& 0.3280 (5.40e-03) 
& 0.2250 (3.20e-02)  
& 0.3120 (7.50e-03) 
& 0.4661 (9.74e-04) 
& \textbf{0.4673 (4.10e-03)}\\
& 100 
& 0.3510 (1.90e-03) 
& 0.3350 (3.80e-03) 
& 0.2450 (2.10e-02) 
& 0.3250 (5.20e-03) 
& 0.4758 (9.14e-04) 
& \textbf{0.4820 (1.80e-03)}\\
\hline
% Continue remaining benchmark problems...

\end{tabular}

\makeatletter
\def\@captype{figure}
\makeatother

\subfigure[$\tau_t=25$]{
    \includegraphics[width=0.31\textwidth]{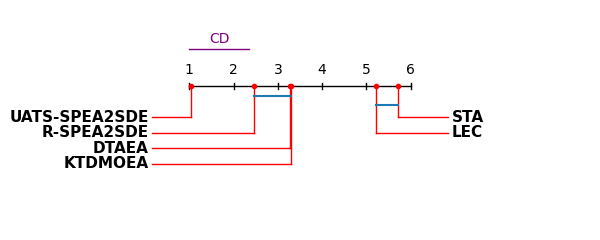}
    \label{fig:setting3_tau25}
}
\hfill
\subfigure[$\tau_t=50$]{
    \includegraphics[width=0.31\textwidth]{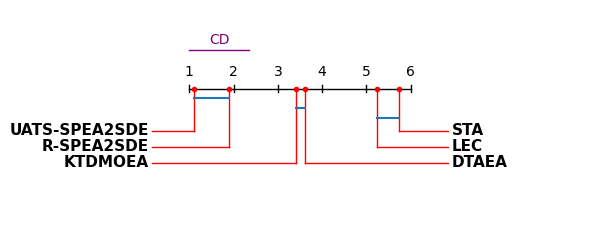}
    \label{fig:setting3_tau50}
}
\hfill
\subfigure[$\tau_t=100$]{
    \includegraphics[width=0.31\textwidth]{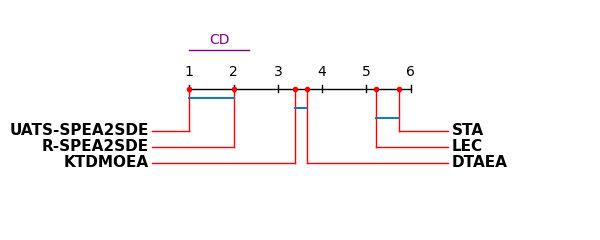}
    \label{fig:setting3_tau100}
}

\caption{Friedman ranking under \textbf{Setting III}. A smaller rank means better performance.}

\label{fig:mhv_setting3}

\end{figure*}

\section{EXPERIMENTS}
This section evaluates UATS on dynamic multi-objective problems with variable objective numbers. The strategy is implemented within SPEA2SDE to obtain UATS-SPEA2SDE, and comparative experiments under several dynamic settings are used to examine convergence and adaptability.

\subsection{Experimental Settings}

\subsubsection{Benchmark Problems}

The experiments adopt a newly proposed benchmark framework \cite{Shang2026Benchmark} and construct four dynamic multi-objective test problems. All benchmark problems are developed based on scalable minus-DTLZ1-4 formulations\cite{Deb2002DTLZ,Shang2026Benchmark} to simulate optimization scenarios with dynamically changing numbers of objectives.

Suppose the complete objective set is defined as

\[
F=\{f_1,f_2,\ldots,f_{10}\}.
\]

At each environmental stage, only a subset of objectives is activated, while inactive objectives are ignored during fitness evaluation. Changing this subset across stages creates different levels of objective-number variation and tests whether an algorithm can adapt and track the PF in a dynamic objective space.

\subsubsection{Dynamic Objective Settings}

To comprehensively evaluate the adaptability of algorithms under different dynamic environments, three objective-changing settings are considered.

\paragraph{Setting I: Mild Objective Change}

Setting I simulates gradual changes in the objective space.
At each change point, one objective is added or removed.
The objective count follows

\[
2 \rightarrow 3 \rightarrow 4 \rightarrow 5 \rightarrow 6
\rightarrow 5 \rightarrow 4 \rightarrow 3 \rightarrow 2.
\]

The active objective subsets are

\[
\begin{aligned}
\{f_2,f_4\}
&\rightarrow
\{f_2,f_4,f_5\}
\rightarrow
\{f_1,f_2,f_4,f_5\}\\
&\rightarrow
\{f_1,f_2,f_4,f_5,f_6\}
\rightarrow
\{f_1,f_2,f_3,f_4,f_5,f_6\}\\
&\rightarrow
\{f_2,f_3,f_4,f_5,f_6\}
\rightarrow
\{f_2,f_3,f_4,f_5\}\\
&\rightarrow
\{f_2,f_3,f_5\}
\rightarrow
\{f_3,f_5\}.
\end{aligned}
\]

This setting tests adaptation to gradual expansion and contraction of the objective space.

\paragraph{Setting II: Moderate Objective Change}

Setting II considers relatively larger objective variations.
Each environmental change adds or removes two objectives.
The objective count follows
\[
2 \rightarrow 4 \rightarrow 6 \rightarrow 8 \rightarrow 10
\rightarrow 8 \rightarrow 6 \rightarrow 4 \rightarrow 2.
\]

The active objective subsets are

\[
\begin{aligned}
\{f_2,f_7\}
&\rightarrow
\{f_2,f_5,f_7,f_{10}\}\\
&\rightarrow
\{f_1,f_2,f_5,f_6,f_7,f_{10}\}\\
&\rightarrow
\{f_1,f_2,f_4,f_5,f_6,f_7,f_9,f_{10}\}\\
&\rightarrow
\{f_1,f_2,f_3,f_4,f_5,f_6,f_7,f_8,f_9,f_{10}\}\\
&\rightarrow
\{f_1,f_2,f_3,f_5,f_6,f_8,f_9,f_{10}\}\\
&\rightarrow
\{f_2,f_3,f_5,f_6,f_9,f_{10}\}\\
&\rightarrow
\{f_2,f_5,f_6,f_9\}\\
&\rightarrow
\{f_5,f_6\}.
\end{aligned}
\]

This setting examines robustness when objective dimensionality changes more strongly.

\paragraph{Setting III: Severe Objective Change}

Setting III simulates highly irregular objective changes.
The objective count follows

\[
2 \rightarrow 5 \rightarrow 10 \rightarrow 6 \rightarrow 3
\rightarrow 8 \rightarrow 4 \rightarrow 7 \rightarrow 9.
\]

The active objective subsets are

\[
\begin{aligned}
\{f_3,f_8\}
&\rightarrow
\{f_2,f_3,f_6,f_7,f_8\}\\
&\rightarrow
\{f_1,f_2,f_3,f_4,f_5,f_6,f_7,f_8,f_9,f_{10}\}\\
&\rightarrow
\{f_1,f_3,f_5,f_6,f_7,f_{10}\}\\
&\rightarrow
\{f_3,f_7,f_8\}\\
&\rightarrow
\{f_1,f_3,f_4,f_5,f_6,f_7,f_8,f_9\}\\
&\rightarrow
\{f_2,f_5,f_7,f_{10}\}\\
&\rightarrow
\{f_1,f_2,f_4,f_5,f_6,f_9,f_{10}\}\\
&\rightarrow
\{f_1,f_2,f_3,f_4,f_5,f_6,f_7,f_8,f_{10}\}.
\end{aligned}
\]

In this setting, some objective-reduction stages are non-nested, meaning that the new active objective subset may contain objectives that were inactive in the previous stage. For these cases, UATS re-evaluates the transferred decision variables instead of applying direct projection.

Compared with Setting I and Setting II, this setting introduces more complex environmental dynamics and is mainly used to evaluate the adaptability and stability of algorithms in highly dynamic optimization scenarios.

\subsubsection{Compared Algorithms}

UATS-SPEA2SDE is compared with a restart-based SPEA2SDE baseline, denoted as R-SPEA2SDE, and four representative dynamic multi-objective optimization algorithms: DTAEA, KTDMOEA, LEC, and STA. Since the original SPEA2SDE is designed for static optimization and does not contain an explicit response mechanism for environmental changes, R-SPEA2SDE reinitializes its population after detecting an objective change and then continues evolution.

The compared algorithms follow the parameter settings recommended in their original studies.

\subsubsection{Parameter Settings}

To ensure fair comparisons, all algorithms use identical population sizes, crossover probabilities, mutation probabilities, and environmental change frequencies.

The shared settings are:

\begin{itemize}
    \item Population size: 300
    \item Environmental change frequencies: $\tau_t = 25, 50, 100$
    \item Maximum generations before the first environmental change: 300
    \item Independent runs: 31
    \item Crossover probability: 1.0
    \item Mutation probability: $1/D$
\end{itemize}
\noindent
where $D$ is the number of decision variables.

For UATS-SPEA2SDE, an unbounded external archive is maintained within each environment stage to store offspring solutions and extract feasible nondominated solutions as transferable elites after environmental changes occur.

\subsubsection{Performance Metrics}

To comprehensively evaluate algorithm performance, the Mean Hypervolume (MHV) metric is adopted in this study.

MHV averages the hypervolume of the obtained solution sets over all environmental stages\cite{While2006HV}. It reflects both convergence and diversity in a dynamic optimization process.

Let $S_t$ be the solution set obtained at time step $t$, and let $T$ be the set of environmental stages. MHV is defined as

\[
MHV = \frac{1}{|T|}\sum_{t \in T} HV(\hat{S}_t),
\]
\noindent
where $\hat{S}_t$ is the normalized objective-vector set at time step $t$.

To reduce the effect of changing objective dimensionality, each hypervolume value is normalized by the hypervolume of the true PF at the corresponding time step.

All reported values are the mean and standard deviation over 31 independent runs.

\subsection{Results and Discussion}

This subsection reports the results under the three dynamic settings. The proposed strategy is incorporated into SPEA2SDE to construct UATS-SPEA2SDE, which is compared with R-SPEA2SDE and four representative dynamic multi-objective optimization algorithms, including DTAEA, KTDMOEA, LEC, and STA.

\subsubsection{Results under Setting I}

Table~\ref{tab:mhv_results_set1} reports the MHV results obtained by the compared algorithms under \textbf{Setting I}.

Experimental results show that UATS-SPEA2SDE achieves superior or competitive performance on most benchmark problems under Setting I, indicating that the proposed transfer strategy improves convergence recovery after mild objective changes.

\subsubsection{Results under Setting II}

Table~\ref{tab:mhv_results_set2} presents the experimental results obtained under \textbf{Setting II}, where the number of objectives changes more significantly. UATS-SPEA2SDE still obtains strong MHV values under larger objective-space variations, demonstrating its robustness to moderate objective changes.

\subsubsection{Results under Setting III}

Setting III introduces highly irregular objective changes and represents the most challenging dynamic environment among all settings. Table~\ref{tab:mhv_results_set3} reports the MHV results of all compared algorithms. UATS-SPEA2SDE remains stable in this setting, which indicates that the proposed transfer mechanism is still effective under severe changes.

Overall, the experimental results demonstrate that the proposed UATS improves environmental adaptability and convergence recovery across different dynamic objective-changing settings.

\section{CONCLUSION}
This paper proposed UATS for dynamic multi-objective optimization with objective-number changes. By maintaining an unbounded external archive within each environment stage, UATS stores offspring solutions and extracts feasible nondominated solutions as transferable elites after environmental changes occur. The proposed strategy was incorporated into SPEA2SDE to construct UATS-SPEA2SDE, aiming to improve convergence recovery and environmental adaptability in dynamic objective spaces.

Experimental studies on four dynamic benchmark problems under three objective-changing settings demonstrated that UATS-SPEA2SDE achieves competitive or superior performance compared with the restart-based SPEA2SDE baseline and four representative dynamic multi-objective optimization algorithms. These results suggest that unbounded archive-guided transfer can accelerate convergence recovery and improve adaptation when objectives are added or removed.

In future work, more refined archive management, transfer selection, and prediction strategies will be investigated to further improve optimization performance in complex dynamic environments.

\section*{Acknowledgement} 
This work was supported by National Natural Science Foundation of China (Grant No. 62472292), Guangdong Basic and Applied Basic Research Foundation (Grant No. 2025A1515011638), Internal Fund of National Engineering Laboratory for Big Data System Computing Technology (Grant No. SZU-BDSC-IF2024-07), and ZTE Industry-University-Institute Cooperation Funds  (Grant No. IA20250707013).
\begingroup
\linespread{0.95}\selectfont

\bibliographystyle{IEEEtran}
\bibliography{references}
\endgroup

\end{document}